# Winning with Less for Low-Resource Languages: Advantage of Cross-Lingual English–Persian Argument Mining Model over LLM Augmentation

ALI JAHAN, Amirkabir University of Technology, Iran
MASOOD GHAYOOMI, Institute for Humanities and Cultural Studies, Iran
ANNETTE HAUTLI-JANISZ, University of Passau, Germany

## Abstract

Argument mining is a subfield of natural language processing to identify and extract the argument components, like premises and conclusions, within a text and to recognize the relations between them. It reveals the logical structure of texts to be used in tasks like knowledge extraction.

This paper aims at utilizing a cross-lingual approach to argument mining for low-resource languages, by constructing three training scenarios. We examine the models on English, as a high-resource language, and Persian, as a low-resource language. To this end, we evaluate the models based on the English Microtext corpus [28], and its parallel Persian translation. The learning scenarios are as follow: (i) zero-shot transfer, where the model is trained solely with the English data, (ii) English-only training enhanced by synthetic examples generated by Large Language Models (LLMs), and (iii) a cross-lingual model that combines the original English data with manually translated Persian sentences. The zero-shot transfer model attains F1 scores of 50.2% on the English test set and 50.7% on the Persian test set. LLM-based augmentation model improves the performance up to 59.2% on English and 69.3% on Persian. The cross-lingual model, trained on both languages but evaluated solely on the Persian test set, surpasses the LLM-based variant, by achieving a F1 of 74.8%. Results indicate that a lightweight cross-lingual blend can outperform considerably the more resource-intensive augmentation pipelines, and it offers a practical pathway for the argument mining task to overcome data resource shortage on low-resource languages.

Authors' Contact Information: Ali Jahan, alijahan@aut.ac.ir, Amirkabir University of Technology, Tehran, Iran; Masood Ghayoomi, M.Ghayoomi@ihcs.ac.ir, Institute for Humanities and Cultural Studies, Tehran, Iran; Annette Hautli-Janisz, Annette.Hautli-Janisz@uni-passau.de, University of Passau, Passau, Germany.

## 1 Introduction

Argument Mining (AM), also known as 'argumentation mining', is a research area between artificial intelligence, natural language processing, and knowledge representation and reasoning which have received researchers' attention in recent years. "Argumentation is a verbal, social, and rational activity aimed at convincing a reasonable critic of the acceptability of a standpoint by putting forward a constellation of propositions justifying or refuting the proposition expressed in the standpoint" [42]. AM aims at detecting the structure of arguments and their relationship from text, including premise, claim, and conclusion, used in various kinds of textual data, such as classroom discussions [24], broadcast political debates [19], presidential campaign debates [16], scientific abstracts [1], student essays [30], and health case [25]. Considering the wide range of applications, having an accurate AM model is very essential for each language.

In terms of the language, however, most of the current works are done on English datasets, such as Stab and Gurevych [39] and Cheng et al. [6]. Other languages have also been studied in this field, but their language resources and datasets are not as wide as English. AM has been studied in the following languages based on the provided datasets: Greek [35], German [9, 22, 28] Italian [3], Chinese [21, 46], Portugues [33], and Spanish [37].

As the globalization demands more inclusive language processing solutions to various problems quickly, including AM, developing different annotated corpora is the first step towards the goals. Annotating such corpora is tedious and costly in terms of time and money, and the development can not happen quickly for all languages across the world. Utilizing cross-lingual models can be a shortcut solution to use the data of a high-resource language, e.g English, for a low-resource language, e.g. Persian. Although significant progress has been made in English-centric AM studies, research on low-resource languages is an active research area.

The current research aims at developing a cross-lingual AM model with English Microtext dataset [28] and evaluating it with its parallel Persian texts, thus contributing to the relatively underexplored area of multilingual argument mining. Our study focuses on two primary tasks: a) argument component detection, and b) relation detection. The former involves identifying Argumentative Discourse Units (ADUs), and classifying their stance as either 'supporting (`pro`)' or 'opposing (`con`)' a central claim. And the latter detects the logical and rhetorical relationships between these units, such as support, rebuttal, undercut, and example.

The imbalance data problem has mostly a counter impact on the performance of a classifier, so that the most frequent label will dominate rare labels. In the current study, a critical challenge is identified during the initial experiments that is the significant unbalancing of the distribution of ADU stance labels, with the '`pro`' label, which overwhelmingly dominates the dataset. To minimize the performance degradation caused by this problem, we employ data augmentation techniques using Large Language Models (LLMs). These models generate high-quality, semantically consistent synthetic ADUs to enhance the minority class, and boost the robustness of the classifier.

The remainder of this paper is organized as follows: After the introduction, Section 2 reviews previous works on AM. In Section 3 a brief overview of the AM task is provided. Section 4 describes the methodology and the three proposed learning scenarios. In Section 5 the detailed information about the datasets, hyper-parameters and evaluation protocol are described. Section 6 reports the experimental results. Section 7, then, offers a qualitative discussion based on two Persian test-set examples, namely LLM-augmented and cross-lingual models. Finally, Section 8 concludes the paper and introduces directions for future work.

## 2 Related Works

Recent advances in AM have increasingly leveraged Transformer-based models to handle complex argumentative structures and enable transfer learning across languages. In the followings, we review the research related to AM.

Habernal and Gurevych [15] created a corpus from web contents, such as comments of articles, discussion forum posts, blog posts, as well as professional newswire articles about six controversial topics in education. With the aim of overpassing the gap between normative argumentation theories and argumentation phenomena encountered in actual data, they adapted an argumentation model tested in an extensive annotation study. They used structural support vector machine with a set of lexical, morphological, sentiment, semantic, and word embedding features to detect argumentative text and classify them as components in the sequence labeling scheme.

Rocha et al. [34] used direct transfer and projection methods for identifying argumentative relations of English and Portuguese. They used the Argumentative Essays corpus [39] and the ArgMine corpus [33]. Experimental results showed that the transfer learning approaches performed better than the in-language approach.

Eger et al. [12] worked on determining whether a sentence includes one of the argumentative structures or not in a sentence classification schema. They used cross-lingual transfer method on the English Student Essay corpus [39] that has also been translated into German. They used a combination of projection and direct transfer methods. They also used TED[1] talks parallel dataset which does not have annotations. Moreover, their approach showed to be robust to change the domain of parallel data even for a domain-specific problem, such as AM.

Eger et al. [11] used machine translation and projection on the source language to create synthesized data for AM in the target language. To this end, they trained the model with a German corpus (the source language) and evaluated the model on English (the target language). They experimented with three datasets: German short texts [28] which is the first parallel English-German AM corpus, argumentative Persuasive Essays [39] automatically translated into German, French, Spanish, and Chinese by Google Translate to achieve a parallel corpus, and Chinese Review Corpus [21] automatically translated into English using Google Translate to develop a parallel corpus. By comparing the model trained with hand-translated data, they observed that the low cost machine-translated synthesized text with projection could almost perform as good as the costly hand-translated one.

Petasis [31] also proposed AM models for Persuasive Essays dataset [39] based on contextual embeddings, such as Bidirectional Encoder Representations from Transformers model (BERT) [8] and Embeddings from Language Model (ELMo) [32] with Bidirectional Long-Short Term Memory model with Conditional Random Field (BiLSTM-CRF). This study examined AM as a fine-grained task and trained deep learning models for sequence labeling problems. They concluded that models without hand engineered features can achieve state-of-the-art results. However, hand engineered features from previous works obtained promising results.

Mayer et al. [25] proposed a complete AM pipeline for health care applications. They created a corpus named AbstRCT[2] by annotating abstracts of randomized controlled trials from the MEDLINE database. In this study, they examined various word embedding models, such as GloVe [29] and FastText [18], and many contextual embeddings including BERT [8], ELMo [32], BioBERT[20] and SciBERT [4]. They used Recurrent Neural Network model, such as gated recurrent units and LSTM or simple dense layers for the sequence tagging task. In addition, the BERT family variants, such as the models for scientific domain and the RoBERTa model [23], were utilized to extract and to classify the relations.

---

[1]https://www.ted.com/
[2]https://gitlab.com/tomaye/abstrct/

Cheng et al. [6] introduced the Integrated Argument Mining corpus, an English dataset annotated with claims, stances, and evidence across multiple debate topics. They proposed two integrated tasks, namely claim extraction with stance classification, and claim–evidence pair extraction. They evaluated both pipelines and end-to-end transformer-based approaches. Results suggested some benefits to joint modeling, although identifying both structure and stance remained difficult, especially across diverse topics.

Sazid and Mercer [36] proposed a unified sequence labeling approach for AM in the Persuasive Essays corpus [39]. Instead of treating component and relation detection separately, their method encodes both into a single prediction task using a transformer-based backbone with additional sequence modeling layers. Experiments on the Persuasive Essays corpus showed competitive results and highlighted the potential benefits of modeling argumentative structure jointly; though, the gains depended on augmentation and architectural choices.

Chen et al. [5] applied AM to analyze cultural variation in English essays written by learners with different language backgrounds. Using transformer-based models trained on existing English data, they segmented argumentative units and compared writing styles across groups of learners with different native languages. Their analysis showed differences in argument organization and explicitness, suggesting potential applications of AM for cross-cultural discourse analysis. However, no new models were introduced, and performance was not the main focus of their research.

Fromm et al. [14] explored argument quality assessment using transformer models under cross-domain and multi-task settings. Their evaluation across multiple datasets revealed generalization issues when models are applied to unseen domains. Incorporating related tasks, such as argument identification, into the training process resulted in slight improvements in certain scenarios. The study highlighted limitations on current argument quality models and the importance of diverse training data.

Segura-Tinoco and Cantador [37] introduced a Spanish corpus for AM in civic forums and studied feature reduction techniques for lightweight modeling. Applying dimensionality reduction to linguistic and embedding features, they found that the reduced representations retain a large portion of classification performance across tasks, such as component classification and relation detection. Their work contributed developing both a new dataset and analyzing efficiency strategies.

Zhao et al. [46] introduced ORCHID[3], a large-scale Chinese corpus of annotated debate transcripts. The dataset includes stance annotations and summaries, and supports a novel task, called stance-specific summarization. Their baselines using pre-trained Chinese transformers indicated that stance classification and summarization in spontaneous debate settings remain challenging. The dataset addressed the lack of resources in Chinese AM and provided a vacant space for future research in non-English languages.

Yeginbergen et al. [45] studied cross-lingual transfer strategies for AM using multilingual transformers. They compared direct transfer model, data translation with annotation projection, and few-shot learning across several languages including Spanish, French, and Italian. Experiments showed that translating data with projected annotations yields more consistent performance than zero-shot transfer. While performance varied across languages, the study emphasized the challenges of cross-lingual stance detection and highlighted the need for language-specific signal even with multilingual models.

Recent progress in computational argumentation is exemplified by an autonomous debating system that can engage competitively with human opponents. Slonim et al. [38] presented a full system architecture and systematic evaluations across diverse topics, including public debates with expert speakers in their research. They argued that debating differs from prior artificial intelligence grand challenges focused on games, since debate requires open-domain retrieval, claim generation, and audience-aware reasoning rather than fixed rules. This perspective motivated new paradigms that integrate language understanding with argumentative planning, aligning with our focus on AM and cross-lingual discourse modeling.

---

[3]Oral Chinese Debate

## 3 Argument Mining

In the introduction section, we briefly over-viewed the major properties of AM. In this section, we provide more detail on this concept. AM roots at argumentation theory which deals with argumentative discourse to analyze the argument's structure and to gain more insight into the textual and contextual pragmatic factors.

Argumentation, as a social activity, aims to either justify and support one's own viewpoint or refute and challenge someone else's. When justifying a viewpoint through argumentation, the goal is to demonstrate that an acceptable proposition is represented. Conversely, when refuting a viewpoint, the aim is to show the idea is unacceptable while its opposite is more reasonable by attacking the viewpoint. Whether justifying or refuting a viewpoint, argumentation involves presenting a series of propositions [43, pp: 3–7], including a) the justifying propositions expressed in the premises, b) the opinion explored in the conclusion or claim, and c) the connection between premises and conclusion.

AM is a computational discourse analysis that automatically discovers the structure of explicit reasoning in text. It includes the following subtasks [41]: a) identifying argumentative text (or a portion of a text); b) segmenting a text into ADUs; c) identifying the central claim; e) identifying the role/function of ADUs; f) identifying relations between ADUs; g) building the overall structural representation; and h) identifying the type and the quality of the argumentation. The proposed models by Palau and Moens [27] and Habernal and Gurevych [15] aimed at identifying the role/function of ADUs. Nguyen and Litman [26] proposed a model for identifying the relationship between ADUs. Moreover, in some researches, such as the proposed model by Eger et al. [10], the argument components and relations were extracted. Ajjour et al. [2] proposed a model to segment the text into ADUs. Ein-Dor et al. [13] proposed a model for retrieving argumentative content.

Early AM studies focused almost exclusively on the English Persuasive Essays corpus [39], legal opinions or user-generated debates, relying on supervised models that demand thousands of carefully annotated training instances. Recent research by Yeginbergen et al. [45], spurred by multilingual transformers, has explored using cross-lingual distributional semantics for AM to transfer knowledge from high-resource languages to low-resource ones.

Distributional semantics is a method for semantic representation of words, phrases, sentences, and documents. This method aims at capturing as much information as possible from the context in a vector space model. It roots at the idea proposed by Wittgenstein [44] who believed that the meaning of a word can be determined by its usage in the language. Harris [17] proposed an idea under the 'distributional hypothesis' such that the words which are used in the same local contexts intend to have a similar meaning. The early study on this domain has focused on monolingual word embedding. Further progress used cross-lingual data to capture the contextual semantic information across different languages. In this paper, we take the advantages of cross-lingual distributional semantics for the AM task to transfer the English knowledge to the Persian data.

## 4 Methodology

### 4.1 Model Architecture

We treat AM as a two-task sequence: (i) ADU stance classification to identify text spans that represent ADUs, and assigns them a stance label either as 'pro' or 'con'; and (ii) relation classification between ADU pairs to predict the presence and type of relations between pairs of ADUs.

To this end, we use XLM-Roberta (XLM-R) [7], a transformer-based multilingual pre-trained language model known for its effectiveness in cross-lingual representation learning. Our model adopts a multi-task architecture in which XLM-R serves as a shared encoder for both argument component detection, and relation detection tasks. On top of the final [CLS] token embedding, two independent linear classification heads are used: i) one for stance classification i.e. 'pro' or 'con'; and ii) another one for relation type prediction, e.g. support, rebuttal, which are described in the following.

- Argument component detection: This task is framed as sentence-level classification at the Elementary Discourse Unit (EDU) level. Each EDU is tokenized and encoded using XLM-R, and the [CLS] embedding is passed through a linear classification head to predict one of two stance labels, namely pro or con.
- Relation detection: For each pair of ADUs with an annotated argumentative link, the model jointly encodes the two ADUs using XLM-R. The [CLS] token embedding of the combined sequence is then passed to a second classification head to predict one of four rhetorical relation types, namely 'support' to indicate that one ADU supports another, 'rebuttal' to contradict or to argue an ADU against another, 'undercut' to weaken or to question the reasoning behind another ADU without directly opposing its claim, or 'example' to connect an ADU to a concrete example or instance that illustrates or supports its argument. The 'segment' relation, which structurally maps EDUs to ADUs, is only used during the preprocessing step, and it is not included in the relation classification task.

### 4.2 Training Setups

The training step for AM is performed using the English Microtext corpus [28], with hyper-parameter tuning via validation splits. The parallel manually translated Persian dataset is used partly for testing. We experiment three training scenarios:

(1) **Zero-shot Transfer Model**: The model is trained with the English Microtext corpus [28] without any changes. In this training phase, no Persian data is used. But the evaluation is performed on both English and Persian test sets separately. The problem with this dataset is the skewness of the data over the pro-dominant class in the dataset. We consider this model as the baseline.

(2) **LLM-Augmented Model**: In this scenario, we try to reduce the imbalance data problem in the zero-shot transfer model. To this end, synthetic English ADUs for the minority con class are generated with LLM models, namely GPT-4[4], Claude[5] 3.7 Sonnet, Gemini[6] 1.5 Flash, and DeepSeek-R1[7]. The generated English examples by the models are appended to the original English training data, yielding a larger but still monolingual training set. The same setups of evaluation with both English and Persian data, as in the previous scenario, are used.

The original English corpus contains 451 pro and 125 con ADUs. For each LLM, we generate 214 additional pro ADUs and 540 con ADUs to balance both classes to the equal 665 instances. All models are accessed through their official APIs with default decoding parameters. The same prompt template is used for all LLMs to ensure a fair comparison; and the only difference is the API endpoint to call the models. All generated ADUs are manually spot-checked to remove malformed outputs before appending to the training set.

---

[4] https://chatgpt.com
[5] https://claude.ai
[6] https://www.gemini.com
[7] https://deep-seek.chat

(3) **En-Fa Cross-lingual Model**: In this learning scenario, we take the advantage of cross-lingual distributional semantics for the AM task. To setup the experiment, the English training set is concatenated with a human-translated Persian version of the data and create a parallel corpus. No additional synthetic data is added. This experiment aims at exploiting cross-lingual properties in the model while keeping the collected data lightweight.

All models are trained using Hugging Face's `Trainer` API with early stopping and `eval_loss` as the tracking metric. The learning rate is fixed at $5 \times 10^{-6}$, and the model is trained for up to 100 epochs. Standard evaluation metrics, namely precision, recall, and F1 scores, along with per-class metrics, are used.

## 4.3 Implementation Details

As described in the previous section, the XLM-R-base is used as a shared encoder for both classification tasks, namely argument component detection and relation detection. The encoder consists of 12 transformer layers, each with multi-head self-attention and feed-forward networks, and a hidden size of 768. Two task-specific linear layers are attached to the encoder: one maps the [CLS] token representation to two stance classes for argument component detection, and the other one maps combined [CLS] embeddings from ADU pairs to one of the five relation types for relation classification.

The classifier heads are implemented as single-layer linear projections followed by softmax activation function. Cross-entropy loss is used for both tasks. Inputs are tokenized using the XLM-R tokenizer; and sequences are padded or truncated to a maximum length of 128 tokens. Each sample includes input_ids, attention_mask, and a label tensor.

The models are implemented in PyTorch and wrapped using Hugging Face's PreTrainedModel interface to ensure compatibility with the trainer API. The AdamW optimizer is used with a learning rate of $5 \times 10^{-6}$, and training is conducted with a batch size of 16, over a maximum 100 epochs with early stopping based on validation loss. Evaluation is conducted at the end of each epoch, and the best model is retained automatically.

## 5 Experimental Setup
## 5.1 Data Set

To train and evaluate the proposed models for the cross-lingual AM task, we use two corpora, namely the the Persuasive Essays corpus [39] and the Microtext corpus [28]. A common property of these datasets is that both of them have used the IOB annotation format.

Stab and Gurevych [39] developed a corpus from Persuasive Essays, annotated with argumentation structures, aiming to improve the accuracy of computational models. They annotated 402 English essays from `essayforum.com`, a community to provide feedback papers and essays. They labeled each argumentative component as 'Major Claim', 'Claim', 'Premise' or 'O'. Stab and Gurevych [40] also defined the type of a relation as 'Support' or 'Attack' and annotated the data.

Peldszus and Stede [28] annotated argument units in 112 German short texts to provide the Microtext corpus. This dataset has one central 'Claim' for each text and several 'Premises'. This corpus does not contain the 'O' label as either a non-ADU or any major claim. This corpus is carefully translated into English to build the German-English dataset. In the current study, this corpus is manually translated into Persian. Tables 1 and 2 reports the statistical information for the Persuasive Essays and Microtext corpora, respectively.

Table 1. Statistical Information of the Persuasive Essays corpus

|                    | EN      |
|--------------------|---------|
| Documents          | 2,235   |
| Sentences          | 8,956   |
| Words              | 148,182 |
| MajorClaim         | 751     |
| Claim              | 1,506   |
| Premise            | 3,832   |
| Major Claims Words | 10,966  |
| Claims Words       | 22,443  |
| Premise Words      | 67,158  |
| NoArg Words        | 47,615  |

Table 2. Statistical Information of the Microtext corpus

|           | EN    | FA    |
|-----------|-------|-------|
| Documents | 112   | 112   |
| Sentences | 453   | 497   |
| Words     | 8,001 | 8,652 |
| pro       | 451   | 451   |
| con       | 125   | 125   |
| pro Words | 6,299 | 6,845 |
| con Words | 1,702 | 1,807 |

The Microtext dataset is structured in the XML format and represents argument graphs hierarchically. Each document is composed of:

- **ADUs**: These are short textual segments annotated with a stance -'`pro`' (proponent) or '`con`' (opponent).
- **EDUs**: Finer-grained text segments that are mapped to ADUs via segmentation links.
- **Relations**: Labeled directed links connecting ADUs, reflecting the argumentative structure of the discourse.

The main challenge in using the Microtext dataset alone is the imbalance labels of ADUs. To overcome this problem, we map labels in the Persuasive-Essays corpus based on the Microtext label scheme. The Persuasive Essays corpus is annotated based on three argument component types, namely *MajorClaim*, *Claim* and *Premise*, together with directed edges labeled *for/against* between a Claim and its MajorClaim, or *support/attack* between a Premise and another component. To convert the Persuasive Essays labels into the two-label scheme in Microtext, we proceed a top-down tree-structured format of the Microtext corpus as follow:

(1) Root mapping: Every *MajorClaim* becomes the tree root and is labeled '`pro`'.
(2) Direct children: For every *Claim* attached to the MajorClaim, we keep the stance information but express it in the Microtext label set: a *Claim* with stance '*for*' is labeled '`pro`', whereas a *Claim* with stance '*against*' is labeled '`con`'.

(3) Recursive premise mapping: For every directed edge originating at a *Premise*:
- If the edge label is *support*, the premise inherits the label ('pro' or 'con') of its target component and the edge itself is stored as support;
- If the edge label is *attack*, the premise receives the *opp* label of its target component and the edge is stored as rebuttal.

The procedure is applied recursively, treating each newly mapped premise as a potential parent; until the entire Persuasive Essays graph is traversed.

The result is a Microtext-style tree whose nodes are labeled pro/con and whose edges are labeled support/rebuttal, and it is fully compatible with the original Microtext annotation.

## 6 Evaluation and Results

In table 3, the results of the different learning scenarios are reported.

### 6.1 Scenario I: Zero-shot Transfer Model

Table 3 (first two rows) presents precision, recall and $F_1$ for the un-augmented zero-shot transfer model for English. In-language evaluation reaches a $F_1$ score of 50.3%; the figure drops only slightly when the same model is tested on Persian (50.8%) because the 'con' instances are rare in both languages. The large precision–recall gap confirms the severe class imbalance pro over con, i.e., more pro than con instances limits the model's ability to retrieve the minority class. This motivates the augmentation and mixing experiments that are described in the following. We treat the obtained results in this experiment as the baseline.

### 6.2 Scenario II: Impact of LLM Augmentation

Table 3 reports the four LLM-augmented models across both evaluation settings using English and Persian data. In the English test, Gemini attains the best $F_1$ score (59.2%), edging out GPT by 2.1%. Once the models are evaluated cross-lingually, GPT takes a clear lead (69.4%), more than eight points ahead of Gemini and over twelve points above Claude.

### 6.3 Scenario III: EN + FA Cross-lingual Model

The cross-lingual setup concatenates the original English corpus with the parallel Persian translation. Table 3 shows that the cross-lingual model raises $F_1$ on the Persian test set to 74.9%, where it outperforms all the models, with or without LLM-augmentations. Comparing the best results, Table 3 emphasizes the benefit of adding genuine Persian data. The cross-lingual model improves $F_1$ score by an absolute 5.5% over the strongest LLM-augmented model (GPT on Persian) and by 24.6% over the baseline, while also delivering the highest precision (0.925). The gains confirm that even a modest amount of human-translated text can outperform considerably more resource-intensive LLM augmentation pipelines in low-resource transfer settings.

## 7 Discussion

To illustrate how the three learning scenarios behave, we analyse two Persian Microtext samples; Case 1 and Case 2 are the files `micro_d14` and `micro_k015` from the test set, respectively. For each text, we show the gold labels for ADUs and relations, along with the predictions of our models. The patterns mirror the aggregate scores reported in Section 6: errors on the minority con label mostly vanish after LLM augmentation, while the cross-lingual model further corrects residual relation mistakes.

Table 3. precision (P), recall (R) and $F_1$ for all scenarios (%)

| Model | | Eval. set | P | R | $F_1$ |
|---|---|---|---|---|---|
| Zero-shot transfer | | EN | 72.4 | 53.1 | 50.3 |
| | | FA | 89.1 | 53.7 | 50.8 |
| LLM-Augmented | GPT | EN | 61.2 | 56.7 | 57.1 |
| | Gemini | | 59.1 | 61.5 | **59.2** |
| | Claude | | 56.3 | 55.2 | 55.4 |
| | DeepSeek | | 63.7 | 51.3 | 46.9 |
| | GPT | FA | 72.3 | 67.8 | **69.4** |
| | Gemini | | 60.7 | 60.9 | 60.8 |
| | Claude | | 38.3 | 48.9 | 42.9 |
| | DeepSeek | | 64.7 | 56.5 | 56.7 |
| EN+FA Cross-lingual | | FA | 92.5 | 70.4 | **74.9** |

Table 4. Per-label precision (P), recall (R) and $F_1$ across all scenarios (%)

| Model | | Eval. set | pro | | | con | | |
|---|---|---|---|---|---|---|---|---|
| | | | P | R | $F_1$ | P | R | $F_1$ |
| Zero-shot transfer | | EN | 0.781 | 98.9 | 87.3 | 66.7 | 7.4 | 13.3 |
| | | FA | 78.3 | 100 | 87.8 | 100 | 7.4 | 13.8 |
| LLM-Augmented | GPT | EN | 79.6 | 91.1 | 85.0 | 42.9 | 0.222 | 0.293 |
| | Gemini | | 83.1 | 71.1 | 76.6 | 35.0 | 51.9 | 41.8 |
| | Claude | | 79.2 | 84.4 | 81.7 | 33.3 | 25.9 | 29.2 |
| | DeepSeek | | 77.4 | 98.9 | 86.8 | 50.0 | 3.7 | 6.9 |
| | GPT | FA | 84.5 | 91.1 | 87.7 | 60.0 | 44.4 | 51.1 |
| | Gemini | | 82.0 | 81.1 | 81.6 | 39.3 | 40.7 | 40.0 |
| | Claude | | 76.5 | 97.8 | 85.9 | 0 | 0 | 0 |
| | DeepSeek | | 79.4 | 94.4 | 86.3 | 50.0 | 18.5 | 27.0 |
| EN+FA Cross-lingual | | FA | **84.9** | 100 | **91.8** | **100** | 40.7 | 57.9 |

**Case 1:**

**EDU1**: *Fa:* من فکر می‌کنم فریتز هرگز در زندگی خود دعوا نکرده است.
*En: I think Fritz has never been in a fight in his life.*

**EDU2**: *Fa:* به عنوان یک پسر، او با سایر پسران کر درگیر می‌شد،
*En: As a boy he did use to scuffle with the other choirboys,*

**EDU3**: *Fa:* اما این به سختی به عنوان یک نزاع مناسب به حساب می‌آید.
*En: but that hardly counts as a proper brawl.*

**EDU4**: *Fa:* و او همیشه وقتی همه چیز خراب می شود، از خود بیخود می شود.
*En: And he always chickens out when things get dicey.*

**EDU5**: *Fa:* دیروز، زمانی که جسور اجازه نداد ما وارد شویم، ناگهان رفت.
*En: Yesterday, when the bouncer wouldn't let us in, he was suddenly gone.*

For relations, the gold graph contains four links: (1) *rebuttal*: ADU2 attacks ADU1; (2) *undercut*: ADU3 weakens the attack link from ADU2 to ADU1 (i.e., it questions the inference, not the nodes themselves); (3) *support*: ADU4 supports ADU1; and (4) *example*: ADU5 provides an example for ADU4.

Table 5. ADU stance predictions for Case 1

| Model | ADU1 | ADU2 | ADU3 | ADU4 | ADU5 |
|---|---|---|---|---|---|
| Gold | pro | **con** | pro | pro | pro |
| Zero-shot Transfer | pro | <u>pro</u> | pro | pro | pro |
| GPT (LLM-aug) | pro | **con** | <u>con</u> | pro | pro |
| EN+FA Cross-lingual | pro | **con** | pro | pro | pro |

The English-only model typically misses the sole con unit (ADU2); and therefore, it predicts an incorrect *support* into ADU1. GPT-based augmentation recovers the con stance for ADU2; but it tends to mistype the undercut as a *rebuttal*. The EN+FA cross-lingual model reproduces the full gold graph, including the *undercut* on the attacking edge and the *example* from ADU5 to ADU4.

**Case 2: TXL_airport_remain_ operational_after_BER_opening**

**EDU1**: *Fa:* BER باید دوباره از ابتدا مفهوم‌سازی شود،
    *En: BER should be re-conceptualized from scratch,*

**EDU2**: *Fa:* حتی اگر میلیاردها یورو قبلاً در پروژه فرودگاه موجود سرمایه گذاری شده باشد:
    *En: even if billions of Euros have already been invested in the existing airport project*

**EDU3**: *Fa:* و این تاریخ تکمیل را برای مدت نامحدودی به تاخیر می‌اندازد.
    *En: and this would delay the date of completion indefinitely.*

**EDU4**: *Fa:* هم عملیات های ساختمانی کشیده شده و هم مسائل ایمنی نصب، کاستی های آشکاری را در کل برنامه ریزی نشان می دهد.
    *En: Both the drawn-out building operations and the mounting safety issues show clear shortcomings in the entire planning.*

Table 6. ADU stance predictions for Case 2

| Model | ADU1 | ADU2 | ADU3 | ADU4 |
|---|---|---|---|---|
| Gold | pro | **con** | **con** | pro |
| Zero-shot Transfer | pro | <u>pro</u> | <u>pro</u> | pro |
| GPT (LLM-aug) | pro | **con** | **con** | pro |
| EN+FA Cross-lingual | pro | **con** | **con** | pro |

In this sample, the gold graph contains three links: (1) *rebuttal*: ADU2 attacks ADU1; (2) *rebuttal*: ADU3 attacks ADU1; and (3) *support*: ADU4 supports ADU1. The English-only model incorrectly treats ADU2 and ADU3 as supportive, producing spurious *support* links. In LLM augmentation (GPT) scenario, both con units are recovered and the two *rebuttal* links are found; and the EN+FA cross-lingual model matches the gold structure.

## 8 Conclusion

This paper set out to probe how far zero-shot transfer, data augmentation with LLMs, and a lightweight cross-lingual model can push up argument-mining performance for genuinely low-resource target languages, like Persian. According to the experimental results, we found that:

- *Zero-shot transfer is feasible but limited.* Training solely on an imbalanced English corpus yields $F_1$ scores of 50.3% (EN → EN) and 50.8% (EN → FA); recall for the minority con class is particularly poor.

- *LLM augmentation helps, especially cross-lingually.* Adding equal numbers of synthetic pro and con ADUs (214 + 540 per class, generated with a uniform prompt through utilizing LLM-based tools, including GPT, Gemini, Claude and DeepSeek) raises $F_1$ to 57.1% in-language and 69.4% cross-lingually, with GPT providing the most transferable cues. All missed con predictions stem from the baseline's exposure to an imbalanced training set. GPT-based augmentation remedies most of them by minority stance recovery.
- *A small human-translated Persian tranche beats all LLM-only variants.* Simply concatenating a manual Persian translation of the English training set lifts $F_1$ score to **0.749** on the Persian test set, an absolute gain of 0.055 over the best LLM model and 0.246 over the baseline, while also delivering the highest precision. Practically, when a handful of accurate translations can be obtained, a cross-lingual models affords larger gains than producing hundreds of synthetic sentences with powerful LLMs, and at a fraction of the computational cost. When LLM augmentation fixes stance labels, relation types can still drift (*suport* vs. *rebuttal*). The cross-lingual model, having seen Persian syntax and discourse markers, assigns relations more accurately.

Promising directions for further studies include (i) enlarging the Persian portion with semi-automatic back-translation, (ii) injecting discourse-level prompts to elicit more diverse minority-class arguments, and (iii) exploring parameter-efficient multilingual fine-tuning to tighten the gap between direct transfer and fully supervised systems.